%% file: main.tex
\documentclass[journal,twoside,web]{ieeecolor}
\usepackage{generic}
\usepackage{amsmath,amssymb,amsfonts}
\usepackage{graphicx}
\usepackage{algorithm}
\usepackage{algorithmicx,algpseudocode}
\usepackage{hyperref}
\hypersetup{hidelinks=true}
\usepackage{textcomp}
\def\BibTeX{{\rm B\kern-.05em{\sc i\kern-.025em b}\kern-.08em
    T\kern-.1667em\lower.7ex\hbox{E}\kern-.125emX}}
\usepackage[maxnames=1, minnames=1, backend=biber, giveninits=true, sorting=none]{biblatex}
\DeclareSourcemap{
    \maps[datatype=bibtex]{
        \map{
            \step[fieldset=ISSN, null]
            \step[fieldset=ISBN, null]
            \step[fieldset=urldate, null]
            \step[fieldset=note, null]
            \step[fieldset=eventtitle, null]
            \step[fieldset=pages, null]
            \step[fieldset=doi, null]
            \step[fieldset=language, null]
        }
    }
}

\newbibmacro*{conditionally-print-doi}{%
  \ifboolexpr{
    test {\iffieldundef{journal}} 
    and 
    test {\iffieldundef{booktitle}}
    and
    test {\iffieldundef{journaltitle}}
  }
  {\printfield{doi}}{}
}

\renewbibmacro*{doi+eprint+url}{
    \usebibmacro{conditionally-print-doi}
}

\usepackage{pdfpages}
\usepackage{tikz}
\usepackage{stfloats}
\usepackage{soul}
\usepackage{subfig}  %
\usepackage{caption} 
\usepackage{multirow}
\usepackage{booktabs}
\definecolor{lightblue}{rgb}{0.8,0.9,1}

\DeclareUnicodeCharacter{2212}{-}

\begin{document}
\title{Personalization on a Budget: Minimally-Labeled Continual Learning for Resource-Efficient Seizure Detection}
\author{Amirhossein Shahbazinia, Jonathan Dan, Jose A. Miranda,  \IEEEmembership{Member, IEEE}, Giovanni Ansaloni, \IEEEmembership{Senior Member, IEEE}, David Atienza, \IEEEmembership{Fellow, IEEE}
\thanks{This research was supported in part by ACCESS – AI Chip Center for Emerging Smart Systems, sponsored by the InnoHK initiative of the Innovation and Technology Commission of the Hong Kong Special Administrative Region Government; the Swiss NSF, grant no. 10.002.812: ”Edge-Companions: Hardware/Software Co-Optimization Toward Energy-Minimal Health Monitoring at the Edge”, and the Wyss Center for Bio and Neuro Engineering with the Lighthouse project on Non-invasive Neuromodulation of Subcortical Structures.}
\thanks{A. Shahbazinia, J. Dan, G. Ansaloni, and D. Atienza are with the Embedded Systems Laboratory, EPFL, 1015 Lausanne, Switzerland (e-mail: amirhossein.shahbazinia@epfl.ch; jonathan.dan@epfl.ch, giovanni.ansaloni@epfl.ch, david.atienza@epfl.ch). }
\thanks{J. A. Miranda is with the Centro de Electrónica Industrial at Universidad Politécnica de Madrid (e-mail: jose.miranda@upm.es).}
\thanks{A. Shahbazinia is the corresponding author. The code is available at \url{https://github.com/Amirho3einS/EpiSMART}.}}

\maketitle

\begin{abstract}
Objective: Epilepsy, a prevalent neurological disease, demands careful diagnosis and continuous care. Seizure detection remains challenging, as current clinical practice relies on expert analysis of electroencephalography, which is a time-consuming process and requires specialized knowledge. 
Addressing this challenge, this paper explores automated epileptic seizure detection using deep learning, focusing on personalized continual learning models that adapt to each patient's unique electroencephalography signal features, which evolve over time.
Methods: In this context, our approach addresses the challenge of integrating new data into existing models without catastrophic forgetting, a common issue in static deep learning models.
We propose EpiSMART, a continual learning framework for seizure detection that uses a size-constrained replay buffer and an informed sample selection strategy to incrementally adapt to patient-specific electroencephalography signals. 
By selectively retaining high-entropy and seizure-predicted samples, our method preserves critical past information while maintaining high performance with minimal memory and computational requirements. 
Results: Validation on the CHB-MIT dataset, shows that EpiSMART achieves a 21\% improvement in the F1 score over a trained baseline without updates in all other patients. On average, EpiSMART requires only 6.46 minutes of labeled data and 6.28 updates per day, making it suitable for real-time deployment in wearable systems. 
Conclusion: EpiSMART enables robust and personalized seizure detection under realistic and resource-constrained conditions by effectively integrating new data into existing models without degrading past knowledge.
Significanc: This framework advances automated seizure detection by providing a continual learning approach that supports patient-specific adaptation and practical deployment in wearable healthcare systems.

\end{abstract}


\input{1-Introduction}

\input{2-Related}

\input{3-Methodology}

\input{4-Experiments}

\input{5-Conclusion}

\printbibliography

\end{document}

%% file: 1-Introduction.tex
\section{Introduction}
\label{sec:intro}

\IEEEPARstart{E}{pilepsy} is a neurological disease that affects millions of people worldwide \cite{gbd_2016_neurology_collaborators_global_2019}, and it manifests itself primarily through the occurrence of seizures.
Seizures are characterized by sudden and unexpected discharges of large groups of neurons, and their detection is crucial for prompt intervention and treatment \cite{fisher_ilae_2014}. The detection of epileptic seizures commonly relies on the analysis of electroencephalography (EEG) signals that capture the electrical activity of the brain. However, manual analysis of long-term EEG signals can be time-consuming and requires expert knowledge. Therefore, there has been growing interest in the development of automated methods for the detection of epileptic seizures.

In particular, deep learning and machine learning techniques, such as convolutional neural networks (CNN), have shown promising results in seizure detection \cite{shoeibi_epileptic_2021, zhou_epileptic_2018, gomez_automatic_2020}.
Unlike traditional machine learning, deep learning models can automatically learn relevant features from raw EEG data, eliminating the need for application-specific feature extraction \cite{pale_exploration_2022}. This ability to learn from raw data makes deep learning particularly suitable for seizure detection, where robust features may not be easily identifiable \cite{abdelhameed_deep_2021, amirshahi_m2d2_2022}.

Deep learning approaches are also well suited to develop personalized models due to their inherent adaptability \cite{baghersalimi_m2skd_2024}. Personalized models adapt to the unique EEG characteristics of each patient \cite{do_handbook_2021}. Considering them can improve performance in the detection of seizures, but training personalized models requires a substantial amount of data \cite{alzubaidi_survey_2023}. These have historically proven costly to collect in a hospital environment. However, the advent of wearable devices now allows long-term data collection \cite{macea_-hospital_2023} and the possibility of fine-tuning existing models rather than training them from scratch. Such continuous, real-world monitoring on wearable devices enables timely seizure detection, even outside of hospital, which can support rapid intervention, improve patient safety, and enhance quality of life.

Personalization introduces its own challenges, particularly due to the non-stationary nature of data collected over time \cite{de_lange_continual_2021}. This is especially relevant in healthcare, where physiological parameters can fluctuate, leading to changes such as electrode impedance, variations in environmental noise, and alterations in exercise routines of patients \cite{do_handbook_2021}.
Deep learning models, even after personalized fine-tuning, are inherently static and, hence, ill-equipped for dynamic environments where data distributions change over time.

Continual learning strategies address this challenge by proposing approaches that enable deep learning models to continuously adapt to new information and environments while avoiding the need to retrain from scratch \cite{parisi_continual_2019}.
The application of continual learning to seizure detection presents specific challenges that require tailored solutions. Seizures are rare events that lead to a highly imbalanced dataset, which complicates the training process. Therefore, it is key to update models according to newly acquired data while at the same time retaining information on past events of high significance (such as seizures), thus avoiding catastrophic forgetting, which is a phenomenon where learning new information causes the model to forget previously acquired knowledge \cite{parisi_continual_2019}.
In addition, personalized healthcare solutions should achieve these goals with the limited resources of mobile devices. 

Recent efforts have explored the use of continual learning for seizure detection \cite{shi_continual_2024, aguilar_metaplastic-eeg_2024}, focusing mainly on adapting models as new data emerge, typically from unseen patients. Although promising, these approaches overlook two critical aspects: the need for personalization and the resource constraints inherent in embedded or wearable systems. A recent study \cite{shahbazinia_resource-efficient_2024} attempts to bridge this gap by incorporating both personalization and resource awareness. However, it assumes full access to the labeled data for each patient and performs model updates on an hourly basis. Such requirements introduce substantial annotation and computational overhead, making the approach impractical for real-world continuous use.

This work addresses these limitations by introducing a continual learning framework for seizure detection at the edge that requires minimal annotation effort, limited memory capacity, and low computational overhead for personalization. Our method incrementally adapts to the evolving EEG patterns of individual patients while preserving crucial information from past seizure episodes. We show that models trained in a Leave-One-Out (LOO) setting fail to generalize effectively to unseen patients, highlighting the necessity of personalization. To overcome this, our approach uses a lightweight replay buffer \cite{chaudhry_tiny_2019} that selectively stores and reuses informative samples from the past. This allows the model to maintain performance over time without incurring prohibitive costs in memory, computation, or labeling.

The main contributions of this paper are as follows:

\begin{enumerate}
    \item We present EpiSMART (Epileptic Seizure Monitoring with Annotation-Reduced Training), a continual learning framework for personalized seizure detection that incrementally adapts to new patient data while preserving past knowledge.

    \item To the best of our knowledge, EpiSMART is the first work which aims at minimizing the amount of required labeled data to enable scalable and memory-efficient continual learning. This design allows the system to operate under labeling, memory, and computation constraints, making it practical for long-term deployment.

    \item We demonstrate that our method is both label and compute efficient. On average, it requires only 6.46 minutes of labeled data and 6.28 updates per day to adaptively update the model, significantly reducing the effort of expert annotation and computational overhead.

    \item We validate our framework in the CHB-MIT dataset, demonstrating performance comparable to fully labeled continual updates, along with a 21.33\% improvement in F1 score and a 77\% reduction in false alarm rates compared to a no-update baseline trained on all other patients and incapable of adapting or mitigating catastrophic forgetting.
\end{enumerate}

The remainder of this paper is organized as follows. In Section \ref{sec:related}, we review related work on seizure detection and continual learning. In Section \ref{sec:method}, we introduce EpiSMART to tackle the challenges of continual learning and update the deep learning model for personalized seizure detection. We evaluate the performance of the proposed method in Section \ref{sec:exp}. Finally, we highlight the main results of the work in Section \ref{sec:conclusion}.

%% file: 2-Related.tex
\section{Related Works}
\label{sec:related}

\subsection{Machine Learning for Epileptic Seizure Detection}

Traditional techniques for detecting epileptic seizures rely on manually extracting features from EEG signals.
These features correspond to both linear and non-linear descriptors of the signal. They are related to the time domain, the frequency domain, the spatial domain (across channels), or their combinations. This process aims at identifying significant features that are manually crafted and then used to train a classifier to differentiate between seizure and non-seizure events \cite{siddiqui_review_2020, paul_various_2018}.

This approach has two limitations. First, it relies on expert knowledge and trial and error when identifying relevant features, leading to a lack of generalizability. Moreover, it is vulnerable to changes in seizure patterns caused by the inherently non-stationary nature of the EEG, which makes its statistical components change across subjects and time \cite{sharma_emerging_2024}.

Deep learning has been extensively used for the automated processing of EEG signals in various contexts, such as
sleep analysis \cite{michielli_cascaded_2019, yildirim_deep_2019},
brain-computer interface \cite{cao_review_2020},
epileptic seizure prediction \cite{khan_focal_2018}
and detection \cite{gomez_automatic_2020, afzal_rest_2024}, demonstrating its ability to learn from raw data and potentially surpass traditional feature extraction-based methods \cite{roy_deep_2019}.

Although traditional deep learning shows remarkable performance, it often assumes that data are independent and identically distributed (i.i.d.). Hence, models developed based on static snapshots of data may not effectively adapt to dynamic, non-stationary environments where this i.i.d. assumption is violated.

\subsection{Continual Learning in Health Care}
\label{sec:related:cl}

Continual learning (CL, also known as incremental learning or life-long learning), is a machine learning approach where a model is sequentially trained on a series of tasks, addressing the challenge of catastrophic forgetting, and thereby
retaining knowledge from previous tasks even when their data are no longer available. This approach is categorized into methods that include the modification of neural architecture (introducing new neurons for new tasks), regularization strategies (controlling the modification of model parameters), and replay buffers \cite{parisi_continual_2019}. 
The latter technique relies on the preservation and use of data points from the past through a buffer in the learning phase and is often seen as the most effective \cite{kiyasseh_clinical_2021}.
Selecting the important samples for the replay buffer remains a challenge, and entropy-based sample selection is widely adopted as an effective strategy \cite{li_entropy-based_2020}.
Research in continual learning has predominantly focused on computer vision \cite{kiyasseh_clinical_2021}. In healthcare, the emphasis has been largely on medical imaging, as seen in \cite{ pianykh_continuous_2020, perkonigg_dynamic_2021, guo_federated_2022, singh_class-incremental_2023}. For instance, \cite{singh_class-incremental_2023} demonstrates the potential of continual learning in general-purpose, shareable AI for medical imaging.

However, continual learning is increasingly being explored to analyze healthcare data series \cite{kiyasseh_clinical_2021, mahmoud_multi-objective_2022, sun_adaptive_2023, li_doctor_2023}. For example, in \cite{li_doctor_2023}, authors propose a multi-disease detection framework using wearable medical sensors and continual learning, which overcomes the limitations of traditional methods by employing a multi-headed neural network and an exemplar-replay-style algorithm for efficient and adaptable disease detection with a single model. Moreover, a continual learning approach with a replay buffer and novel trainable task-specific parameters is designed in \cite{kiyasseh_clinical_2021} to address performance degradation in deep learning systems due to non-i.i.d. clinical data.

The use of continual learning in EEG-based applications has also recently gained attention due to the inherent non-stationarity and subject variability of EEG signals \cite{zhou_brainuicl_2024, li_domain-incremental_2024, duan_online_2024, mei_ultra-low_2024}. For example, \cite{zhou_brainuicl_2024} introduces BrainUICL, a framework for individual continual unsupervised learning in various EEG tasks, enabling robust adaptation to new users in real-world scenarios. Similarly, \cite{li_domain-incremental_2024} proposes a domain-incremental learning model for motor imagery classification that mitigates forgetting through adversarial training and memory replay. In \cite{duan_online_2024}, a memory-based continual learning approach is designed for streaming EEG scenarios. Therefore, it handles subject imbalance and identity uncertainty through adaptive memory selection and online shift detection. Complementing these, \cite{mei_ultra-low_2024} demonstrates an embedded continual learning solution in a wearable BMI system, showing notable gains in accuracy and energy efficiency during inter-session adaptation.

\begin{figure*}[t]
    \centering
    \includegraphics[scale=0.5]{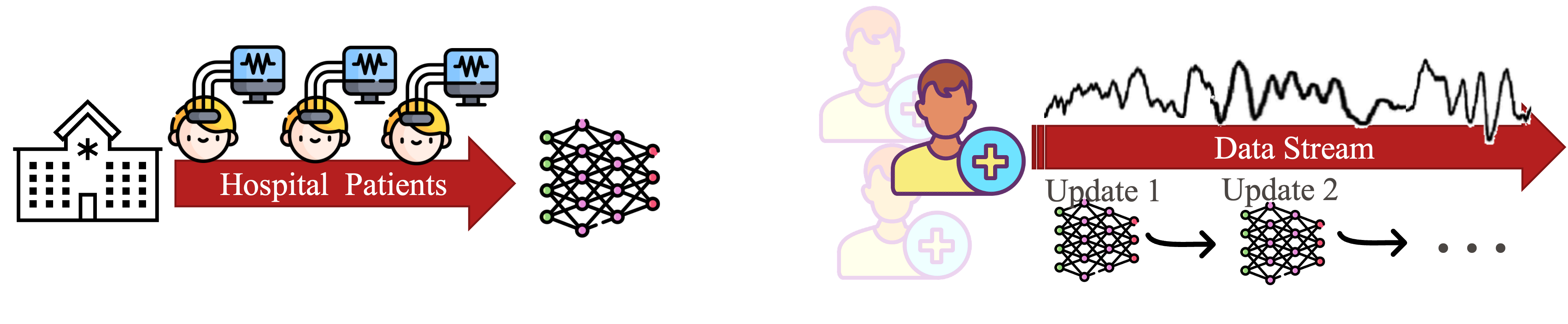}
    \caption{Initially, a subject-independent model is trained using collected data (left). The model is then fine-tuned and personalized to an individual patient by incorporating information from his/her own recordings as they become available over time (right).}
    \label{fig:meth:cl}
\end{figure*}

Continual learning has also been explored in the context of seizure prediction and detection. Shi et al. \cite{shi_continual_2024} propose a Memory Projection strategy which, along with regularization and dimensionality reduction techniques, incrementally learns EEG patterns from patients within dynamically defined subspaces. Similarly, Aguilar et al. \cite{aguilar_metaplastic-eeg_2024} introduce a biologically inspired approach using metaplasticity to improve the stability of binarized neural networks when exposed to sequential EEG datasets. In addition, Shahbazinia et al. \cite{shahbazinia_resource-efficient_2024} propose a replay-based continual learning strategy for personalized seizure detection. Their method emphasizes resource efficiency and demonstrates that with only a small replay buffer and minimal updates, the model can achieve performance on par with a fully supervised, resource-unlimited approach while substantially reducing false alarms.

Despite these advances, most existing works require extensive labels and frequent updates, which limits their practical deployment. Conversely, our work builds upon these efforts by addressing three key challenges in personalized seizure detection: the non-stationarity of EEG data, the strong class imbalance due to the rarity of seizures, and the computational limitations of real-world embedded health monitoring systems. By doing so, we offer a continual learning framework that is not only effective but also practical for long-term, real-time seizure monitoring in personalized settings.

%% file: 3-Methodology.tex
\section{Methodology}
\label{sec:method}

\subsection{Overview and Problem Description}

As depicted in Fig. \ref{fig:meth:cl}, our framework for continual learning in seizure detection considers as input a subject-independent model. This is derived by training a model on a database containing signals from a large number of subjects.
This initial model is the starting point for personalized updates, which are made progressively as new data are acquired from a patient. Therefore, our methodology enables the patient-specific refinement of the initial model to suit the unique and changing physiological signals of each individual. It achieves this goal while a) coping with a memory constraint (which must be taken into account on personal health monitors), b) retaining information on relevant past data to avoid catastrophic forgetting, and c) minimizing the need for labeled data.

We consider a streaming setup in which EEG recordings arrive sequentially over time. Formally, the data stream is represented as \( \mathcal{X} = \{x_t \mid x_t \in \mathbb{R}^{n} \} \) for each timestamp \( t \), where each sample \( x_t \) consists of EEG readings across \( n \) channels. The stream \(\mathcal{X}\) serves two purposes: a) it is used in real-time for inference and seizure detection, and b) it supports personalization by enabling sample selection and model updates. 
As new samples are received, the model selects and stores high-entropy samples, along with predicted seizures, in a memory buffer. Data are kept in a size-constrained buffer and progressively evicted as new samples arrive. When the number of retained samples reaches a predefined threshold, an update is triggered, enabling the model to refine its parameters. 
In this scenario, the EpiSmart methodology, described in the following,  enables the personalization process while minimizing the effort of labeling, memory usage, and computational cost.

\subsection{EpiSMART}
\label{sec:meth:alg}

\begin{figure*}[t]
    \centering
    \includegraphics[scale=0.48]{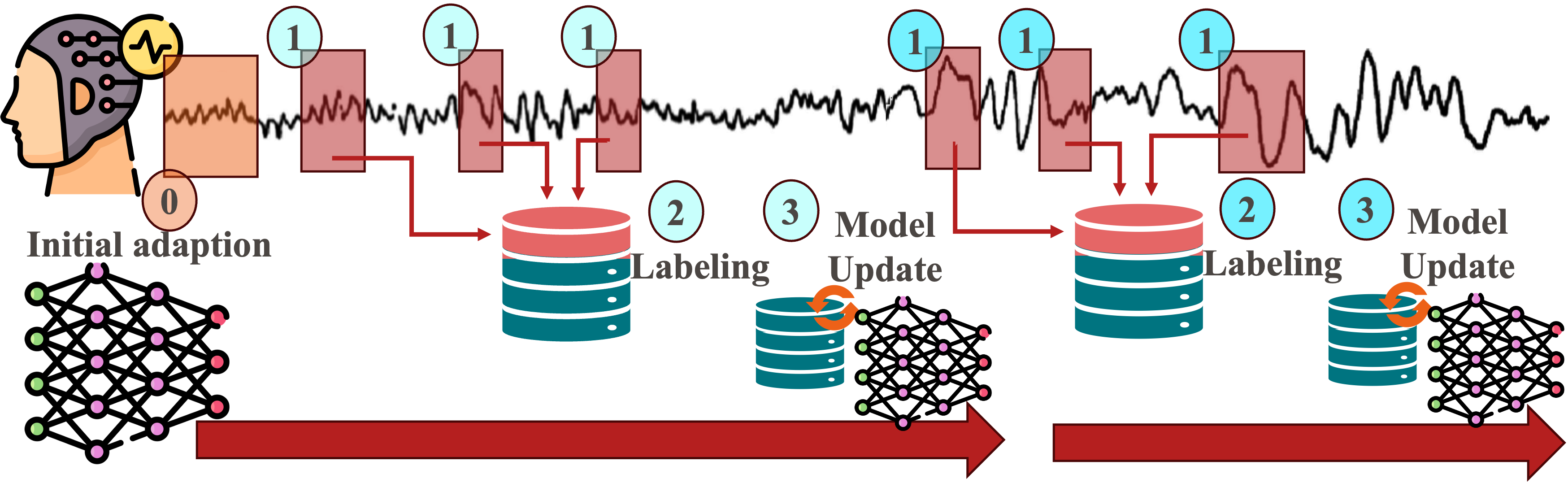}
    \caption{Overview of the EpiSMART framework. The method follows a four-stage process: (0) Initial adaptation using the first available labeled data, (1) Selection of important samples based on entropy and predicted seizures, (2) Labeling of newly selected samples, and (3) Model fine-tuning using the labeled data and past representative samples stored in the buffer.}
    \label{fig:meth:alg}
\end{figure*}

Our approach is illustrated in Fig. \ref{fig:meth:alg}, which presents the high-level view for continual seizure detection and adaptation. The methodology consists of four key stages: 0) an initial adaptation phase, followed by iterative steps for 1) sample selection, 2) labeling, and 3) model updates.

In deployment, \textbf{0) initial adaptation} is performed using the first available labeled data. This phase establishes a personalized model by refining the subject-independent model \( M \) with non-seizure and seizure data from the individual. The fine-tuned model is then used for real-time inference as new EEG data are acquired. However, this one-time adaptation does not account for the non-stationarity of EEG signals over time, and relying on large amounts of labeled data for adaptation is impractical in real-world scenarios.

The continual learning process begins with \textbf{1) selecting important samples} from the incoming data stream. At each time step \( t \), an entropy-based selection mechanism identifies uncertain predictions, storing samples with entropy values above a predefined threshold \( \tau_E \). High-entropy samples are those for which the model’s confidence is low, with predicted probabilities for seizure and non-seizure being close to each other. Such ambiguous cases are particularly valuable for improving the model (as shown in Section \ref{sec:exp:random}). Furthermore, samples for which the model predicts a seizure event (\( \hat{y}_t = 1 \)) are retained, ensuring that rare but critical events are captured in the memory buffer \( B \).

To determine high-entropy samples, the entropy of each new sample \( x_t \) is computed using the model's predicted probability distribution. Given the unnormalized logits (i.e., the model’s output before the softmax layer) \( \ell_i \), the probability distribution over the classes is obtained using the softmax function:

\begin{equation}
    p_i = \frac{\exp(\ell_i)}{\sum_{j} \exp(\ell_j)} , \quad i \in \{0,1\}
\end{equation}

where \( p_i \) represents the probability of class \( i \). The entropy \( H(x_t) \) of this distribution quantifies the uncertainty of the model and is calculated as follows:

\begin{equation}
    H(x_t) = - \sum_{i\in \{0,1\}} p_i \log p_i
\end{equation}

Furthermore, the predicted label \( \hat{y}_t \) is determined by selecting the class with the highest probability.

\begin{equation}
    \hat{y}_t = \arg\max_{i \in \{0,1\}} p_i
\end{equation}

Once enough samples are accumulated, the next step is \textbf{2) labeling important samples}. When the number of high-entropy and classified seizure samples reaches the update threshold \( \tau_U \), expert annotation is required for newly acquired samples. This ensures that only the most informative data are labeled, reducing annotation costs and maintaining model performance. In fact, the number of required labeled data is minimal (see Section \ref{sec:exp}).

Following labeling, the \textbf{3) model update} phase is initiated. The model undergoes fine-tuning using the labeled samples and the existing buffer, which retains a representative set of past data to prevent catastrophic forgetting.

The framework is formulated in Algorithm \ref{alg:adaptive_update}. Personalization begins with an initial adaptation step (line 3), where the model \( M \) is fine-tuned using the initial labeled data. The adaptation process is performed iteratively. In Stage 1 (lines 6–11), important samples are selected. Stage 2 (line 13) is activated when all newly acquired data have been selected. Finally, in Stage 3 (line 14), the model is fine-tuned with the updated data.

\begin{algorithm}
\caption{EpiSMART: Epileptic Seizure Monitoring with Annotation-Reduced Training}
\label{alg:adaptive_update}
\begin{algorithmic}[1]
\State \textsc{Initialize}: Pre-trained model $M$, entropy threshold $\tau_E$, update threshold $\tau_U$, memory buffer $B$, sample counter $C$
\State $B \gets$ Initial labeled data
\State Fine-tune $M$ using data in $B$
\State $C \gets 0$

\For{each incoming EEG sample $x_t$ at time $t$}
    \State Compute entropy $H(x_t)$
    \State Predict label $\hat{y}_t$ using $M$
    \If{$H(x_t) > \tau_E$ \textbf{or} $\hat{y}_t = 1$}
        \State Store $x_t$ in $B$ and update $B$ (discard old samples)
        \State $C \gets C + 1$
    \EndIf
    \If{$C \geq \tau_U$}
        \State Label newly added samples in $B$
        \State Fine-tune $M$ using data in $B$
        \State $C \gets 0$
    \EndIf
\EndFor

\end{algorithmic}
\end{algorithm}

\subsection{Memory Buffer}
\label{sec:meth:mb}

The buffer is structured to maintain a balanced representation of seizure and non-seizure samples, ensuring that both classes contribute effectively to model adaptation. To achieve this, the buffer is equally divided between seizure and non-seizure data. However, since seizure events are rare, there may not always be enough seizure samples to fill the allocated partition. In such cases, the remaining space is temporarily assigned to non-seizure data, ensuring memory utilization without wasting available capacity.

Our method adapts continuously over time by incorporating a sample replacement strategy that adjusts to the incoming data stream.
Unlike the approach in \cite{shahbazinia_resource-efficient_2024}, which requires two separate memory to store historical and newly acquired data, our method integrates both new and historical data within a unified buffer, thereby reducing storage demands.
As new samples are added, either due to high entropy or predicted seizures, older samples are randomly evicted to maintain a fixed buffer size. Since older samples are replaced at random, there is always a chance for them to survive; thus, the buffer maintains a diverse representation of past events without favoring specific time periods.
An exception is seizure samples, which are 
prioritized because of their clinical importance. If seizure occurrences are underrepresented, all available seizure samples are retained.
This buffering approach ensures that the model retains exposure to both recent and historical patterns, which is essential for generalization and robust adaptation while learning from the more frequent non-seizure events.

%% file: 4-Experiments.tex
\section{Experiments}
\label{sec:exp}

\subsection{Experimental Setup}
\label{sec:exp:setup}

\subsubsection{Dataset Overview and Data Preparation}
\label{sec:exp:setup:data}
The database used in this study is the CHB-MIT dataset \cite{shoeb_application_2009, goldberger_physiobank_2000}. This public dataset comprises 982 hours of EEG recordings from 24 pediatric patients suffering from intractable epilepsy.
Out of the 982 hours of recordings, only three hours contain seizure events, representing a total of 198 seizures. This dataset is one of the largest public datasets in terms of hours of recording per patient, with an average of approximately 40 hours per patient. 
The EEG signals are sampled at 256 Hz with a 16-bit resolution. Most, but not all, records feature 23 bipolar channels with electrodes placed according to the International 10–20 system. Here, for the sake of uniformity and comparability, we have considered only the 18 channels present in all patients. Three filters were applied to preprocess the data: a 0.5 Hz highpass, a 60 Hz lowpass, and a 50 Hz notch. All filters are 4th-order Butterworth filters. 

We considered data arriving in a continuous stream, structuring the EEG recordings so that each patient's data stream begins with a seizure event. This setup ensures that personalization on seizure data can be performed from the initial labeled data. Following the approach in \cite{shahbazinia_resource-efficient_2024}, only the first hour of the data is used for the initial adaptation. Similarly, we employ a replay buffer capable of storing one hour of data.

Data augmentation was employed on seizure windows in training sets, both when deriving the initial patient-independent model and for patient-dependent fine-tuning steps, in order to equalize the number of seizure and non-seizure data. We used a combination of time shifting (shifting consecutive windows by 1/8 of a second to generate new data windows) and data repetition.

\subsubsection{Implementation}
Our methodology is agnostic with respect to the structure of the deep neural network used. Without loss of generality, in all experiments, we employed a domain-specific fully convolutional network (FCN) tailored for EEG analysis introduced in \cite{gomez_automatic_2020}.
The FCN takes input windows of 4 seconds and predicts a binary label corresponding to a seizure or non-seizure window. It comprises three blocks, each consisting of a convolution layer, batch normalization, a Rectified Linear Unit (ReLU) activation function, and a pooling layer. This is followed by two fully convolutional layers and a concluding SoftMax layer. 
The network has low resource requirements, employing approximately 300K parameters, and hence is a representative choice in the context of personalized health monitoring.

The training procedure is consistent across all cases, running for \(100\) epochs with early stopping applied using patience of \(15\) epochs. The optimization process utilizes cross-entropy loss and the Adam optimizer, initialized with a learning rate of \(1e-4\). To dynamically adapt the learning rate, we employ a strategy in which the learning rate is reduced by a factor of \(10\) if the validation loss does not improve for \(10\) consecutive epochs, avoiding performance plateaus during training. All experiments are repeated three times with different random seeds to ensure robustness. Subject-independent models were trained only once due to their high computational cost.

\subsubsection{Post Processing}

The model is designed to predict a label for each 4-second window. These windows overlap by 3 seconds, providing continuity in data analysis. Seizure events exhibit temporal dependencies, as seizure samples are often correlated over time and are not random. 
Post-processing improves detection performance by enforcing these temporal dependencies, taking into account the context of seizure events over time. Therefore, first, a simple moving average filter is applied to smooth out the predictions. Subsequently, for a more accurate evaluation, event-based scoring is adopted as in \cite{dan_szcore_2024}. This method assesses the performance of the model at the level of seizure episodes instead of using sample-by-sample-based metrics, ensuring that the evaluation reflects clinically relevant outcomes.

\subsubsection{Metrics}
\label{exp:metrics}

In our experiments, we evaluated seizure detection performance using event-based F1 score and False Alarm Rate (FAR). The F1 score, calculated as the harmonic mean of precision and recall, assesses the model's ability to correctly identify seizures while minimizing false detections. It is formally defined as:

\begin{equation}
\text{Precision} = \frac{TP}{TP + FP}, \quad
\text{Recall} = \frac{TP}{TP + FN}
\end{equation}

\begin{equation}
\text{F1 score} = 2 \times \frac{\text{Precision} \times \text{Recall}}{\text{Precision} + \text{Recall}}
\end{equation}
where \(TP\) (true positives) represents correctly detected seizures,  \(FP\) (false positives) indicates non-seizure events mistakenly classified as seizures, and \(FN\) (false negatives) are missed seizures.

The False Alarm Rate (FAR) measures the number of false alarms per day and is computed as:

\begin{equation}
\text{FAR} = \frac{\text{Total False Alarms}}{\text{Days of Recording}}
\end{equation}

These metrics are computed on a stream, which means that each sample is evaluated using the most recent version of the model.

Beyond performance evaluation, we introduce two additional metrics to quantify the cost of each method. The first is labeling cost, a crucial consideration given the significant expense associated with obtaining expert-annotated data. This metric is measured in minutes per day, representing the average amount of data that must be labeled for each patient:

\begin{equation}
\text{Labeling Cost} = \frac{\text{Total Labeled Data (minutes)}}{\text{Days of Recording}}
\end{equation}

The second cost metric is the update overhead, which is particularly relevant for wearable applications where resource efficiency is critical. We quantify this by counting the number of model updates performed per day. Since each update is performed using a fixed-size replay buffer and the number of epochs per update remains relatively consistent, the total number of updates is linearly related to the update cost:

\begin{equation}
\text{Update Cost} = \frac{\text{Total Updates}}{\text{Days of Recording}}
\end{equation}

Across the experiments, the first hour of labeled data (initial adaptation), required for all experiments, is excluded from both labeling and update cost metrics.

\subsubsection{\textit{EpiSMART} Hyper-parameters and Baseline Methods}
\label{exp:baseline_methods}

When not otherwise specified, we evaluate \textit{EpiSMART} with an entropy threshold \( \tau_E \) = \(1e-5\) and an update threshold \( \tau_U \) = \(15\). The different hyper-parameter choices are explored in Section \ref{sec:exp:hp}. 

We compare \textit{EpiSMART} with the following baselines:

\begin{enumerate}
    \item \textit{ No Update}: This method utilizes a subject-independent pre-trained model trained on data from other patients, without any personalized fine-tuning, as in \cite{gomez_automatic_2020}. We employ a Leave-One-Out setting in which the model is trained on all data from the 23 other patients.
    \item \textit{Update Every Hour}: This baseline updates the model every hour, as in \cite{shahbazinia_resource-efficient_2024}, taking into account changes in the patient’s EEG over time through frequent fine-tuning.

\end{enumerate}

\subsection{Experimental Results}
\label{sec:exp:results}

In this section, the experimental results of our proposed method are presented. 
We report the performance and cost of the proposed method with the evaluation metrics introduced in Section \ref{exp:metrics}. 

\subsubsection{Cost/Performance Analysis}

\begin{figure}[t]
    \centering
    \resizebox{\columnwidth}{!}{
    \includegraphics[scale=0.6]{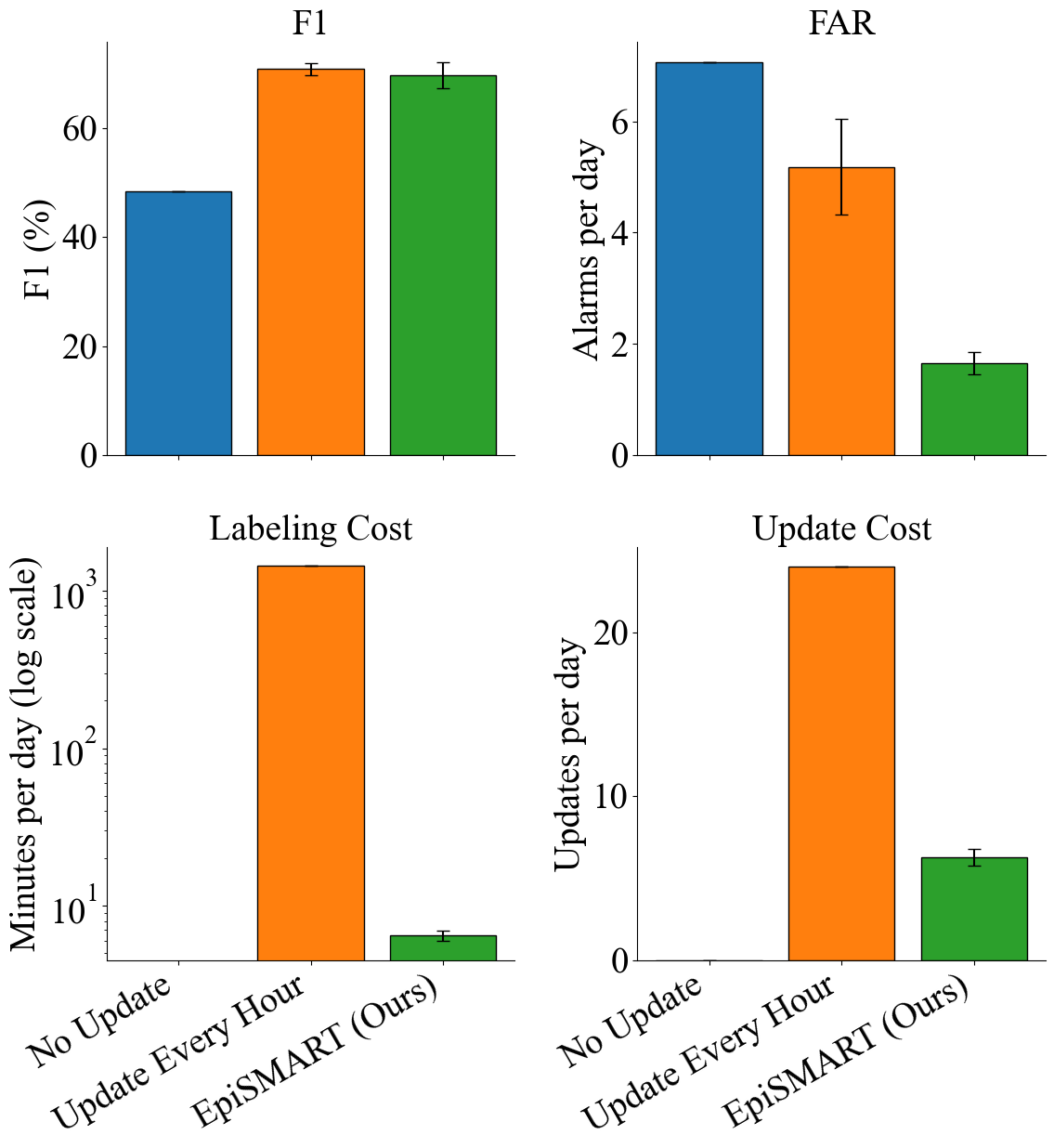}
    }
    \caption{F1 score, FAR, Label and Update Cost, for the three considered methodologies: No Update, Update Every Hour, and EpiSMART. Higher values for F1 score, and lower values for FAR, labeling cost, and update cost indicate better performance.}
    \label{fig:exp:general}
\end{figure}

Fig. \ref{fig:exp:general} comparatively evaluates the cost and performance of \textit{EpiSMART} with respect to the two baseline methods. The top row reports the F1 scores and False Alarm Rates (FARs), while the bottom row highlights the associated labeling and update costs.

The \textit{No Update} baseline achieves an F1 score of \(48.34\) and a FAR of \(7.07\). This method requires neither data labeling (after the initial training) nor model updates. However, its notably low performance shows the limitations of subject-independent models in practical applications, emphasizing the necessity of personalized adaptation.

The \textit{Update Every Hour} method yields an F1 score of \(70.76 \pm 1.16\) and a FAR of \(5.18 \pm 0.86\). The results highlight that its frequent fine-tuning when compared to \textit{EpiSMART}, each employing limited data, is not always beneficial and can even degrade performance by increasing the number of false alarms. Moreover, despite the substantial improvement in detection performance with respect to \textit{No Update}, the associated costs are significant. In fact, this approach requires the labeling of all data for each patient, resulting in a significant annotation burden that is unrealistic in real world settings due to the reliance on expert input.  In addition, the model must be updated every hour, which imposes a high update cost, particularly for wearable devices where efficiency is critical.

\textit{EpiSMART} achieves an F1 score of \(69.69 \pm 2.42\) with an FAR of \(1.65 \pm 0.2\), demonstrating performance comparable to the hourly update strategy. However, the associated costs are considerably reduced.
Unlike \textit{Update Every Hour}, which retains a separate memory for new data, our method relies solely on a unified historical memory buffer, effectively halving the memory requirement (see Section \ref{sec:meth:mb}).
Moreover, instead of requiring all data to be labeled, the proposed approach requires only an average of \(6.46\) minutes of labeled data per day. Furthermore, the update cost is significantly lowered, reducing the number of updates from \(24\) per day to just \(6.28\). This substantial reduction in labeling, update, and memory costs highlights the efficiency of the proposed method while maintaining competitive seizure detection performance.

\subsubsection{Hyper-parameter Exploration} 
\label{sec:exp:hp}
\begin{figure*}[t]
    \centering
    \includegraphics[scale=0.22]{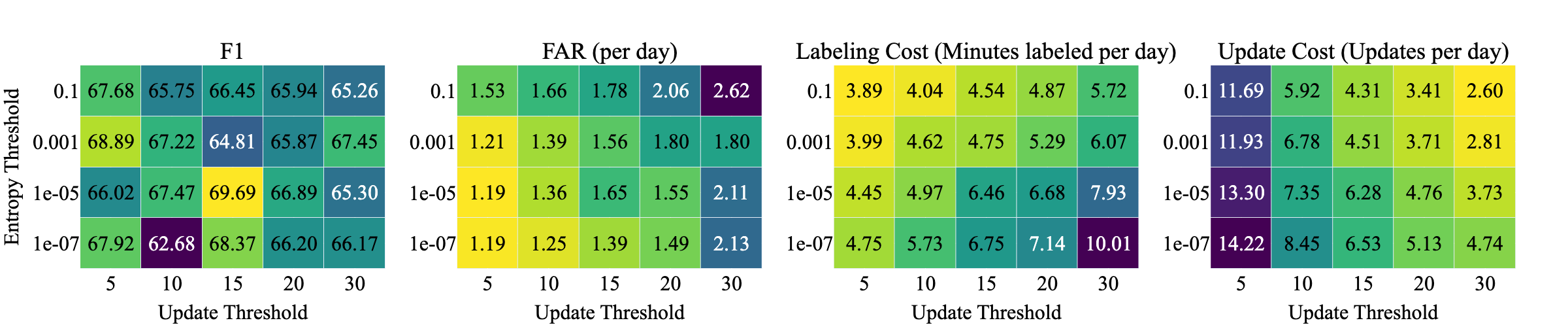}
    \caption{Visualization of metrics across varying entropy thresholds ($\tau_E$) and update thresholds ($\tau_U$). Brighter colors indicate better performance—higher values for F1 score, and lower values for FAR, labeling cost, and update cost.}
    \label{fig:exp:heat_map}
\end{figure*}

Fig. \ref{fig:exp:heat_map} presents heatmaps of various metrics across different entropy and update thresholds. Brighter colors indicate better results: higher values for the F1 score and lower values for FAR, labeling, and update costs. As discussed in Section \ref{sec:meth:alg}, the entropy threshold determines how high a sample's uncertainty must be for it to be retained, while the update threshold controls the frequency of updates based on the accumulated number of samples. These two parameters significantly influence the overall performance and cost of the method. This is visualized in the first and second subfigures of Fig. \ref{fig:exp:heat_map}, which show how different threshold values impact the F1 scores and FARs. 

The third and fourth subfigures from the left illustrate the labeling and update costs, respectively. The cost of labeling is minimized when fewer samples are selected using a higher entropy threshold. In addition, a lower update threshold also results in lower labeling costs. This observation suggests that when the model performs poorly, it generates a larger proportion of high-entropy samples, and delayed updates lead to more of these samples accumulating. Consequently, the most cost-effective strategy for labeling is to set both a high entropy threshold and a high update threshold to ensure more frequent updates.

Regarding the update cost, the primary influencing factor is the update threshold, where higher values reduce the number of updates required. The entropy threshold also influences this process: Higher thresholds result in fewer updates, which is reasonable as they decrease the number of selected samples.

The first and second subfigures from the left demonstrate the performance of different hyperparameter settings. The FAR metric exhibits a more consistent dependence on hyper-parameters compared to the F1 score. This discrepancy arises because seizure events are rare and updating the model does not necessarily introduce new seizure information, making it challenging to observe a consistent pattern in the F1 scores. 
As shown in the second subfigure, more frequent updates result in lower FAR values.
This suggests that increasing computational (update) effort can help reduce FAR, although it may also lead to higher labeling costs.

In conclusion, setting the update and entropy thresholds too low or too high leads to increased update or labeling costs. These parameters vary across patients, and there are no universally optimal values. However, moderate values for both thresholds tend to yield the best overall performance. For example, an entropy threshold of \(1e-5\) combined with an update threshold of \(15\), as we adopted elsewhere in this section, yields strong average performance across metrics.

\subsection{Ablation Study}
\label{sec:exp:ablation}

\subsubsection{Selecting Random Samples}
\label{sec:exp:random}
This section explores the impact of different sampling strategies within the proposed methodology. As discussed in Section \ref{sec:method}, two key criteria govern sample selection and replay buffer updates: selecting high-entropy samples based on a defined entropy threshold and incorporating samples that the model predicts as seizures. Two experimental scenarios were investigated to assess the significance of these criteria, which we compared to \textit{No Update} and \textit{EpiSMART} strategies.

The first scenario examines whether entropy-based selection is essential by instead selecting samples randomly. This ensures a fair comparison, as the sampling follows a uniform distribution to maintain the same labeling and update costs as the proposed method, averaging 6.5 minutes of labeled data and 6.3 updates per day. The second scenario builds upon the first by incorporating seizure-labeled samples to analyze whether including such instances enhances performance. These  are named \textit{Random Update (RU)} and \textit{Random Update + Seizure Labeled (RU+SL)} in
Table \ref{tab:exp:ablation:sample}.

\begin{table}[t]
\caption{Impact of sampling strategies on model performance and update Costs. \(\uparrow\) indicates higher is better, \(\downarrow\) indicates lower is better.}
\centering
\resizebox{\columnwidth}{!}{
\begin{tabular}{|c|c|c|c|c|}
    \hline
    \multirow{2}{*}{Experiment} & \multicolumn{2}{c|}{Performance} & \multicolumn{2}{c|}{Cost} \\
    \cline{2-5} & F1 \(\uparrow\) & FAR \(\downarrow\) & Labeling  \(\downarrow\) & Update \(\downarrow\) \\ \hline
    No Update       & \(48.36\) & \(7.07 \) & \(0\) & \(0\)\\ \hline
    RU     & \(48.99\) & \(20.83\) & \(6.62\) & \(6.3\)\\ \hline  
    RU + SL       & \(51.59\) & \(15.49\) & \(42.3\) & \(5.61\)\\ \hline  
    \textbf{EpiSMART}           & \(\mathbf{69.69}\) & \(\mathbf{1.65}\) & \(\mathbf{6.46}\) & \(\mathbf{6.28}\)\\ \hline  
\end{tabular}
}
\label{tab:exp:ablation:sample}
\end{table}

As observed in the table, \textit{RU} performs poorly. The primary reason is that randomly selected samples do not contribute meaningful information and progressively replace valuable data in the replay buffer, leading to performance degradation over time. Including seizure-labeled samples (\textit{RU + SL}) slightly improves performance but at a considerable labeling cost. This is due to the model's declining accuracy over time, which leads to an increasing number of falsely predicted seizure samples entering the training process, worsening the performance degradation.

In conclusion, effective sample selection is crucial to maintain model performance. Randomly selecting samples leads to performance deterioration, and while incorporating seizure-labeled samples offers some improvement, it is not cost-effective. The proposed method, which selects high-entropy samples while balancing labeling and update costs, achieves superior performance with significantly lower resource requirements.

\subsubsection{Neighborhood data}

In this section of the ablation study, we investigate the impact of incorporating neighborhood samples. The rationale behind this approach is that adding neighboring samples of a high-entropy event may help the model better distinguish patterns and learn richer feature representations. Clinically, doctors also evaluate samples in the context of their surrounding data rather than relying solely on isolated windows \cite{do_handbook_2021}. 

\begin{figure}[t]
    \centering
    \includegraphics[scale=0.18]{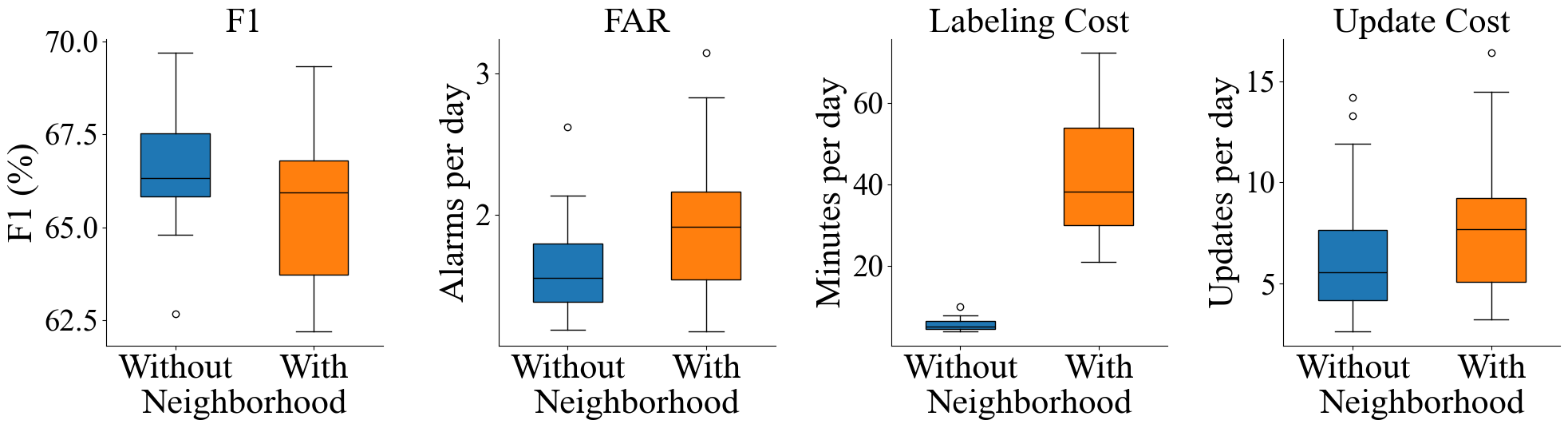}
    \caption{Impact of adding neighborhood data on performance and cost metrics across multiple hyperparameter settings.}
    \label{fig:exp:ablation:neigh}
\end{figure}

For this experiment, whenever a sample is selected based on its entropy value or predicted label, its neighboring samples are also labeled and added to the memory buffer. 
These additional neighboring samples do not contribute to the update count. In this setting, we compare the results to a scenario in which one minute of neighborhood data is added along with each selected sample.
Fig. \ref{fig:exp:ablation:neigh} presents the results with and without neighborhood data. The metric distributions shown are obtained by repeating the experiment—under all hyperparameter configurations (see Section \ref{sec:exp:hp})—three times with different random seeds. The subfigures, from left to right, show the F1 score, FAR, Labeling Cost, and Update Cost.

Regarding performance, the F1 score tends to decrease slightly, as indicated by the median values of the experiments, while FAR shows a marginal increase. This suggests that placing more data in the buffer does not necessarily provide additional useful information compared to retaining old samples.

On the cost side, the Labeling Cost increases dramatically with the addition of one-minute neighborhood data, rising from a median of \(5.13\) minutes to \(38.22\) minutes. In addition, the inclusion of neighborhood data increases the update cost. This increase is attributed to the model updates that degrade performance over time, leading to an accumulation of high-entropy samples and more frequent model updates.

Incorporating neighborhood samples does not provide significant performance benefits and drastically increases labeling costs. Therefore, selecting high-entropy samples without their neighbors remains the most efficient approach for maintaining performance while minimizing cost.

%% file: 5-Conclusion.tex
\section{Conclusions}
\label{sec:conclusion}

Physiological signals vary from patient to patient and tend to change over time, even for the same patient. These characteristics pose a challenge for their automated analysis.

In this paper, we have tackled it from multiple perspectives. Focusing on the concrete scenario of seizure detection from EEG data,  we first showed that deep learning models can be effectively fine-tuned on a patient-specific basis, taking into account novel information as it is acquired. 
Next, we introduced a strategy to retain historical data over the long term using a size-constrained replay buffer. Finally, we showed that a very small amount of novel information is sufficient to adapt models while mitigating forgetting, and that this can be automatically selected.

Experimental results show that our subsequent EpiSMART methodology improves performance compared to a strong Leave-One-Out baseline that does not perform model personalization, achieving a 21.33\% gain in F1 score. Moreover, the method requires only 6.46 minutes of labeled data and 6.28 updates per day on average, while relying on a compact 1-hour replay buffer.